\pdfoutput=1
\documentclass[11pt]{article}

\usepackage[]{acl}

\usepackage{times}
\usepackage{latexsym}

\usepackage[T1]{fontenc}
\usepackage[utf8]{inputenc}

\usepackage{microtype}

\usepackage{tabularx}
\usepackage{bm}
\usepackage{amsmath}
\usepackage{mathrsfs}
\usepackage{multirow}
\usepackage{booktabs}
\usepackage{tikz}
\usepackage{tikz-qtree}
\usepackage{color}
\usepackage{xcolor}
\usepackage{makecell}
\usepackage[switch]{lineno}
\usepackage{enumitem}
\usepackage{makecell}
\usepackage{bbding}
\usepackage{ulem}
\usepackage{graphicx} 
\usepackage{subcaption}

\title{An Interpretability Evaluation Benchmark for Pre-trained Language Models}

\author{
Yaozong Shen,
Lijie Wang,
Ying Chen,
Xinyan Xiao,
Jing Liu,
Hua Wu \\
Baidu Inc, Beijing, China \\
\{shenyaozong,wanglijie@baidu.com\}\\
}

\begin{document}
\maketitle

\begin{abstract}

While pre-trained language models (LMs) have brought great improvements in many NLP tasks, there is increasing attention to explore capabilities of LMs and interpret their predictions. 
However, existing works usually focus only on a certain capability with some downstream tasks. 
There is a lack of datasets for directly evaluating the masked word prediction performance and the interpretability of pre-trained LMs. 
To fill in the gap, we propose a novel evaluation benchmark providing with both English and Chinese annotated data.
% to comprehensively evaluate prediction ability and interpretability of pre-trained LMs.
It tests LMs abilities in multiple dimensions, i.e., grammar, semantics, knowledge, reasoning and computation. 
In addition, it provides carefully annotated token-level rationales that satisfy sufficiency and compactness. 
It contains perturbed instances for each original instance, so as to use the rationale consistency under perturbations as the metric for faithfulness, a perspective of interpretability.
We conduct experiments on several widely-used pre-trained LMs. The results show that they perform very poorly on the dimensions of knowledge and computation. And their plausibility in all dimensions is far from satisfactory, especially when the rationale is short. In addition, the pre-trained LMs we evaluated are not robust on syntax-aware data.
We will release this evaluation benchmark at \url{http://xyz}, and hope it can facilitate the research progress of pre-trained LMs.
\end{abstract}

% keywords can be removed
%\keywords{Pre-trained Language Models \and Model Interpretability Evaluation \and Ability of Language Models}

\section{Introduction}

\begin{table*}[tb]
\small
\renewcommand\tabcolsep{2.5pt}
\centering
\scalebox{0.9}{
\begin{tabular}{l | c | c | c}
\toprule
\multirow{2}{*}{Dimensions} & \multirow{2}{*}{Collected Instances} & \multicolumn{2}{c}{Masked Inputs and Golden Answers}\\
\cline{3-4}
 & & Masked Input & Golden Answer \\
\hline
Grammar & \makecell[l]{Aeroflot 's international fleet of 285 \textbf{planes} is being \\ repainted and refurbished at Shannon Airport.} & \makecell[l]{Aeroflot 's \textcolor{red}{international fleet of} \textcolor{red}{285} [MASK] is being \\ repainted and refurbished \textcolor{red}{at} Shannon \textcolor{red}{Airport}.} & planes\\
\hline
Semantics & \makecell[l]{The city of Austin has a total \textbf{area} of 703.95 \\ square kilometres.} & \makecell[l]{The \textcolor{red}{city} of Austin \textcolor{red}{has} a total [MASK] of 703.95 \\  \textcolor{red}{square kilometres}.} & area \\
\hline
Knowledge & \makecell[l]{The country \textbf{Germany} is located directly to the east \\ of Belgium.} & \makecell[l]{The country [MASK] is \textcolor{red}{located directly to the east} \\ \textcolor{red}{of Belgium}.} & Germany \\
\hline
Reasoning & \makecell[l]{The man's eyes were stabbed by broken glass, \\ then he went \textbf{blind}.} & \makecell[l]{The man's \textcolor{red}{eyes were stabbed} by broken glass, \\ then he went [MASK].} & blind \\
\hline
Computation & \makecell[l]{Tony planted a 4 foot tree. The tree grows at a rate of \\ 5 feet every year. It takes \textbf{5} years to be 29 feet.} & \makecell[l]{Tony planted a \textcolor{red}{4 foot tree}. The tree \textcolor{red}{grows} at a rate of \\ \textcolor{red}{5 feet every year}. It takes [MASK] \textcolor{red}{years} to \textcolor{red}{be 29 feet}.} & 5\\
\bottomrule
\end{tabular}
}
\caption{Examples for five evaluation dimensions. The words in the bold are masked and then predicted by LMs. The words in the red color are taken as the ground-truth rationale for answer prediction.}
\label{tab:example}
\end{table*}

\iffalse
\begin{table*}[tb]
\renewcommand\tabcolsep{2.5pt}
\centering
\begin{tabular}{l | l }
\toprule
Dimensionality & Instance \\
\hline
Grammar & \makecell[l]{Aeroflot 's international fleet of 285 \textcolor{red}{planes} is being repainted and refurbished at Shannon Airport.} \\
Semantic & The City of Austin has a total \textcolor{red}{area} of 703.95 square kilometres. \\
Knowledge & The country \textcolor{red}{Germany} is located directly east of Belgium.\\
Reasoning & The man's eyes were stabbed by broken glass, then he went \textcolor{red}{blind}.\\
Computation & \makecell[l]{Tony planted a 4 foot tree. The tree grows at a rate of 5 feet every year. It takes \textcolor{red}{5} years to be 29 feet.}\\
\bottomrule
\end{tabular}
\caption{Examples for five dimensionality. The words in the red color will be masked and then predicted by the model.}
\label{tab:example}
\end{table*} 
\fi
Pre-trained LMs such as BERT \cite{devlin2019bert} and RoBERTa \cite{liu2019roberta} have achieved significant gains in predictive accuracy on a variety of NLP tasks \cite{wang2018glue}.
Many studies have proved that pre-trained LMs have learned amounts of knowledge from the massive text corpora (i.e., their training data), such as linguistic knowledge \cite{tenney2019bert, jawahar2019does} and factual knowledge \cite{petroni2019language, poerner2019bert}.
Such learned knowledge has enhanced representations and capabilities of LMs, e.g., the abilities of reasoning \cite{brown2020language} and computation \cite{polu2020generative}.
However, some works show that pre-trained LMs have not captured adequate knowledge and are insufficient in some aspects.
Some studies find that BERT has not learned some syntactic structures and can not perform well on syntax-aware data \cite{wang2019tree, min2020syntactic}.
Some works state that pre-trained LMs have a poor grasp of reasoning over factual knowledge and commonsense \cite{poerner2019bert, marcus2020gpt}.
Meanwhile, some researchers prove that pre-trained LMs have a poor performance on mathematical problem solving, even on simple problems \cite{hendrycks2021measuring, cobbe2021training}.
Consequently, what kind of knowledge is learned and to what degree it is learned by pre-trained LMs are still unclear. Meanwhile, there is a lack of datasets for comprehensively evaluating model capabilities.

On the other hand, interpreting the decision-mechanism of a pre-trained LM which can help us understand the reason behind its success and its limitations has attracted lots of attention \cite{rethmeier2020tx, meng2022locating, geva2022lmdebugger}. 
With input saliency methods \cite{smilkov2017smoothgrad, sundararajan2017axiomatic}, \citet{ding2021evaluating} use the most influential tokens in the context as the rationale and evaluate interpretability from the perspective of grammar.
Some works \cite{voita2019bottom, singh-etal-2019-bert} study the inner workings of transformer-based pre-trained LMs according to hidden states and evolutions of hidden states between layers.
In addition, some researchers develop toolkits to capture, analyze and visualize inner mechanisms of LMs at the level of individual neurons \cite{rethmeier2020tx, dai2021knowledge, alammar-2021-ecco, geva2022lmdebugger}.
However, most of these studies lack quantitative evaluation and analysis.

To address the above problems, we propose a novel evaluation benchmark for pre-trained LMs, which contains instances with masked words and corresponding human-annotated rationales. As shown in Table \ref{tab:example}, the masked words are used to evaluate model predictions, and the rationales are used to evaluate interpretability.
Overall, our contribution includes:
\begin{enumerate}[leftmargin=*]
\item To our best knowledge, this is the first benchmark that can be used to evaluate both prediction performance and interpretability of pre-trained LMs. And it provides both English and Chinese evaluation sets.
\item Our evaluation benchmark covers common evaluation dimensions, i.e., grammar, semantics, knowledge, reasoning and computation. We create perturbed instances and evaluate faithfulness via the consistency of rationales under perturbations.
%\yaozong{这一点和第一点是不是同一个东西？需不需要加可解释方面的结论？}
%\lijie{我们的亮点要都写出来，可以放在一起，但有一些不突出；可解释的结论在第三条.这部分没有标红，都看一下吧}
\item We conduct experiments on several widely-used pre-trained LMs, such as BERT, RoBERTa. The experimental results show that current pre-trained LMs have very poor prediction performance on the dimensions of knowledge, reasoning and computation. And these LMs are less robust on syntactically transformed data. We believe these findings can help improve LMs.
\end{enumerate}
\section{Related Work}
\label{sec:related_work}

%In this section, we first review studies on variants of pre-trained LMs and analyses about their capabilities. 
In this section, we first review analyses about pre-trained LMs' capabilities. 
Then we introduce studies on LMs' interpretation, including interpretation methods (i.e., rationale extraction methods), evaluation datasets and metrics. 

\subsection{Capability Analyses of Pre-trained LMs}
\label{sub-sec:capability analysis}

%Pre-trained LMs such as BERT \cite{devlin2019bert} have rapidly improved the state-of-the-art on many NLP tasks \cite{wang2018glue}. 
%According to model structure, pre-trained LMs can be divided into three categories: 1) encoder models, including BERT \cite{devlin2019bert}, RoBERTa \cite{liu2019roberta}, ERNIE \cite{sun2019ernie}, and XLM \cite{lample2019cross}; 2) decoder models, such as GPT family \cite{radford2018gpt, radford2018gpt2, Brown2020gpt3}; 3) encoder-decoder models, e.g., T5 \cite{raffel2019T5} and BART \cite{lewis2019BART}.

While such pre-trained LMs have attracted lots of attention and been developed rapidly, it remains unclear why they work well or fail on some inputs, which limits further hypothesis-driven improvement of the architecture.
Consequently, a large number of studies attempt to reveal the reasons behind their performance \cite{jawahar2019does, hewitt2019structural, poerner2019bert}.

%\paragraph{Capability Analysis of Pre-trained Language Models}
%grammar and semantic
Many studies aim to unveil linguistic structures from the representations of pre-trained LMs to understand what kind of linguistic knowledge they have learned \cite{jawahar2019does, kim2020pre}.
\citet{hewitt2019structural} learn a linear transformation to predict the syntactic depth of each word based on its representation, and state that syntax information is implicitly embedded in BERT.
\citet{jawahar2019does} and \citet{tenney2019bert} show that BERT captures rich linguistic information, with syntactic features at lower layers and semantic features at higher layers.

%knowledge
Meanwhile, some studies intend to analyze model performance on capturing factual knowledge.
Based on BERT's good performance on answering cloze-style questions about relational facts between entities, \citet{petroni2019language} state that BERT memorizes factual knowledge during pre-training.
However, \citet{poerner2019bert} prove that the impressive performance of BERT is partly due to reasoning about the surface form of entity names. They filter out queries that are easy to answer from entity names alone, and show that BERT’s precision drops dramatically.
%They conduct experiments by filtering out queries that are easy to answer from entity names alone, and show that BERT’s precision drops dramatically.

%reasoning
Similarly, some studies try to verify the reasoning ability of pre-trained LMs. 
GPT-3 \cite{brown2020language}, a powerful language generator, is proved that it can generate samples of news articles which human evaluators have difficulty distinguishing from articles written by humans.
But \citet{marcus2020gpt} state that GPT-3 has no idea what it's talking about, and show that GPT-3 has a poor grasp of reasoning over commonsense.

%computation
Finally, some studies focus on analyzing the computation capability of pre-trained LMs.
\citet{polu2020generative} propose GPT-f, an automated prover and proof assistant, which applies transformer-based LMs to automated theorem proving by generating the Metamath formalization language.
\citet{hendrycks2021measuring} release an evaluation dataset MATH and a large auxiliary pre-training dataset AMPS to measure the capabilities of machine learning models on mathematical problem solving. Their results show that even those enormous transformer models such as GPT-3 \cite{brown2020language} get relatively low accuracy.
\citet{cobbe2021training} publish the dataset GSM8K, which contains 85,000 high-quality linguistically diverse grade school math word problems, to evaluate model performance on multi-step mathematical reasoning. They find that even the largest transformer models fail to achieve high test performance, while a bright middle school student should be able to solve every problem.

\subsection{Interpretation of Pre-trained LMs}

\paragraph{Interpretation Methods}
%主要分为三类：来自输入，神经元，隐层表示
Over the recent years, interpreting predictions of LMs has attracted immense attention.
The existing studies mainly interpret model predictions from three perspectives: input saliency \cite{ding2021evaluating}, hidden state \cite{alammar-2021-ecco, geva2022lmdebugger} and neuron activation \cite{rethmeier2020tx, meng2022locating, alammar-2021-ecco, dai2021knowledge}.
Input saliency methods assign an importance score to each input token, which represents the token's impact on model prediction \cite{simonyan2014deep, smilkov2017smoothgrad, sundararajan2017axiomatic, li2016visualizing}. \citet{ding2021evaluating} use different saliency methods to interpret LMs on two proposed tasks, i.e., subject-verb number agreement and pronoun-antecedent gender agreement, and evaluate their performance with extracted rationales.
Hidden states and their evolution between model layers are always used to glean information about the inner workings of an LM.
\citet{singh-etal-2019-bert} utilize hidden states to study internal representations of multilingual BERT.
\citet{voita2019bottom} use hidden states to analyze the flow of information inside transformers, and reveal how the representations of individual tokens and the structure of the learned feature space evolve between layers across different tasks.
Examination of neuron activations is used to trace and analyze model processes by extracting underlying patterns of neuron firings.
\citet{rethmeier2020tx} modify the established computer vision explainability principle of ``visualizing preferred inputs of neurons'' \cite{erhan2009visualizing} to suit NLP, and use it to quantify knowledge changes or transfers during training at the level of individual neurons.
\citet{dai2021knowledge} introduce the concept of knowledge neurons and propose a knowledge attribution method to identify the neurons that express the fact.
Meanwhile, several tools are released to capture, analyze, visualize, and interactively explore inner mechanisms of LMs \cite{dalvi2019neurox, alammar-2021-ecco, geva2022lmdebugger}.

Our work focuses on interpreting model predictions with input saliency methods, i.e, post-hoc explanation methods, among which four types are commonly used, namely attention based, gradient based, eraser based, and linear based methods.
Attention based methods use attention weights as the importance scores \cite{jain2019attention, wiegreffe2019attention, Cui2022multi}. 
Gradient based methods use model back propagation gradients as the importance scores \cite{simonyan2014deep, smilkov2017smoothgrad, sundararajan2017axiomatic, li2016visualizing}. 
In eraser based methods, the token importance score is measured by the change of model prediction confidence when the token is removed \cite{li2016understanding, lundberg2017unified, feng2018pathologies}.
Linear based methods use an explainable linear model to approximate the evaluated model behavior locally and use the learned token weights as importance scores \cite{ribeiro2016should, alvarez2017causal}.
Considering the characteristics of LMs, attention based methods and gradient based methods are often used as saliency methods for LMs.

\paragraph{Evaluation Datasets}
Recently, many evaluation datasets with human-annotated rationales are published to facilitate the research progress of interpretability \cite{deyoung2020eraser, wang2021dutrust, camburu2018snli}.
The rationales in such datasets are mainly presented in three forms: highlights \cite{deyoung2020eraser, mathew2021hatexplain}, structural rules \cite{camburu2018snli, rajani2019explain}, and free texts \cite{ye2020teaching, geva2021did}.
While attention to construct interpretability evaluation datasets for specific NLP tasks keeps increasing, there is a lack of evaluation datasets for LMs.
\citet{ding2021evaluating} create four datasets for two tasks, i.e., number agreement of a verb with its subject and gender agreement of a pronoun with its related mentions. They provide token-level annotations in the four sets, each with a cue set and a corresponding attractor set, and use them to evaluate interpretability of saliency methods.
\citet{akyurek2022tracing} propose an fact tracing dataset with instance-level rationales to evaluate model capability on fact learning. 
\citet{he2022pretrained} design a novel task called Simile Property Probing to test whether pre-trained LMs can interpret similes or not by letting the LMs infer the shared properties of similes.

%In order to comprehensively evaluate capabilities and interpretability of pre-trained LMs, we provide a novel evaluation benchmark with token-level rationales.

\paragraph{Evaluation Metrics}
Plausibility and faithfulness are often used to evaluate interpretability \cite{doshi2017towards, jacovi2020towards, adebayo2020Debugging}.
Plausibility measures how much the rationales provided by models align with human-annotated rationales \cite{weerts2019human, deyoung2020eraser, mathew2021hatexplain}. 
With rationales at different granularity levels, several metrics are proposed to measure plausibility, such as token F1-Score \cite{mathew2021hatexplain, wang2021dutrust}, IOU (Intersection-Over-Union) F1-Score and AUPRC (Area Under the Precision-Recall curve) score \cite{deyoung2020eraser}. 
Faithfulness measures the degree to which the rationales in fact influence the corresponding predictions \cite{jacovi2020towards, deyoung2020eraser, liu2020interpretations}. Similarly, multiple metrics are proposed to evaluate faithfulness, e.g. sufficiency and comprehensiveness \cite{deyoung2020eraser}, consistency under perturbations \cite{ding2021evaluating, wang2021dutrust}, sensitivity and stability \cite{yin2021on}, as well as word-level differential reaction \cite{mosca2022that}.

In our work, we also evaluate interpretability from perspectives of plausibility and faithfulness. And we adopt metrics that are suitable for LMs, as described in Section \ref{sec:metrics}.
%For faithfulness evaluation, we adopt the consistency of rationales under perturbations as metrics, as it is suitable for LMs.

\section{Evaluation Dataset Construction}
\label{sec:data_construction}

Our evaluation dataset is constructed in three steps: 1) data collection; 2) perturbed data construction; 3) iterative rationale annotation and checking. 
We first introduce our proposed five evaluation dimensions in Section \ref{ssec:eval_dimension}.
Then we describe the annotation process in Section \ref{ssec:data_collection}-\ref{ssec:disturb_data}. 
Finally, we give our data statistics. 
%Meanwhile, we show other annotation details in Appendix \ref{sec:data_other}.

\subsection{Evaluation Dimensions}
\label{ssec:eval_dimension}

Considering the abilities that an LM should have for predicting the right answer, we design five evaluation datasets with five types of instances, as described below. The corresponding examples are shown in Table \ref{tab:example}.
\begin{itemize}[leftmargin=*]
\item \textbf{Grammar}. These instances are designed to evaluate what linguistic knowledge a pre-trained LM has learned, such as the tense of a verb, the gender of a pronoun, and the number of a noun in English. As shown by the first example in Table \ref{tab:example}, the noun right after the number ``285'' must be plural if it is countable. 
\item \textbf{Semantics}. These instances aim to test whether the pre-trained LM has learned and mastered lexical meaning, including conceptual senses of words, concept properties and relationships, as well as semantic coreference rules. The second example in Table \ref{tab:example} requires the model to master the concept of ``city'' and its property ``area''.
\item \textbf{Knowledge}. The instances in this type are used to evaluate the extent to which the pre-trained LM learns real-world factual knowledge. As shown by the third example in Table \ref{tab:example}, the prediction of ``Germany'' requires the model to learn and memorized related knowledge.
\item \textbf{Reasoning}. These instances aim at the inferential capability of LMs over open-domain commonsense. The forth example in Table \ref{tab:example} states that the model should deduce ``blind'' according to the premise that the eye was hurt.
\item \textbf{Computation}. The instances are intended to test the computational ability of pre-trained LMs on handling mathematical problems, as illustrated by the last example in Table \ref{tab:example}.
\end{itemize}

\subsection{Data Collection}
\label{ssec:data_collection}

\begin{table*}[tb]
\renewcommand\tabcolsep{2.5pt}
\centering
\scalebox{0.62}{
\begin{tabular}{l | c | c }
\toprule
Dimension & English & Chinese \\
\hline
Grammar & Penn Treebank-3\footnote{\url{https://catalog.ldc.upenn.edu/LDC99T42}} & Chinese Treebank 8.0\footnote{\url{https://catalog.ldc.upenn.edu/LDC2013T21}}, Chinese Dependency Treebank 1.0\footnote{\url{https://catalog.ldc.upenn.edu/LDC2012T05}}\\
Semantics & Wikipedia\footnote{\url{https://huggingface.co/datasets/wikipedia}}, WebNLG \cite{moryossef2019step}, WSC \cite{levesque2012winograd} & Baidu Baike\footnote{\url{https://baike.baidu.com}}, DuIE \cite{li2019duie}, CLUEWSC2020\footnote{https://github.com/CLUEbenchmark/CLUEWSC2020} \cite{xu2020clue}\\
Knowledge & FreebaseQA \cite{jiang2019freebaseqa} & CKBQA\footnote{\url{https://github.com/pkumod/CKBQA}, the dataset for knowledge-based question answering task in CCKS 2019.}\\
Reasoning & COPA \cite{roemmele2011choice} & XCOPA \cite{ponti2020xcopa} \\
Computation & Alg514 \cite{kushman2014learning}, Dolphin18K \cite{huang2016well} & Math23K \cite{wang2017deep} \\
\bottomrule
\end{tabular}
}
\caption{Datasets used to create our datasets. For datasets from LDC, we have been authorized.}
\label{tab:raw_data}
\end{table*} 
In order to create high-quality evaluation datasets, we construct our datasets on the basis of some existing human-annotated datasets, as shown in Table \ref{tab:raw_data}. 
Our collection process consists of two steps: instance construction and masked word selection. 

\paragraph{Instance Construction}
Each input in our evaluation datasets is a sentence or a paragraph consisting of multiple sentences, as shown in Table \ref{tab:example}. Since the input forms of some existing datasets in Table \ref{tab:raw_data} are not as we want them to be in our datasets, we have them manually modified. Specially, we create inputs for dimensions of \textit{knowledge}, \textit{reasoning} and \textit{computation}.

\textbf{Knowledge}. We build our English evaluation dataset based on FreebaseQA \cite{jiang2019freebaseqa}, and build the Chinese dataset on CKBQA. An original input includes two required components: a question and its answer. We replace the wh-phrase in the question with the corresponding answer to form a new input. For example, the forth example in Table \ref{tab:example} was originally ``\textit{Which country is located directly to the east of Belgium?}'' with the answer ``Germany''. We use ``Germany'' to replace the wh-phrase ``Which country'' and construct a new input ``Germany is located directly to the east of Belgium.''.
We filter out questions with multiple answers to ensure the uniqueness of the prediction.

\textbf{Reasoning}. We select COPA \cite{roemmele2011choice} and XCOPA \cite{ponti2020xcopa} to build English and Chinese evaluation data, respectively. Each original input has three parts: a given premise and two plausible alternatives for either the cause or the effect of the premise. We concatenate the premise and the reasonable cause to build a new input using some appropriate conjunctions, such as ``since'' and ``because''. Similarly, we use conjunctions such as ``then'' and ``so'' to connect the premise and the proper effect to create a new input.   

\textbf{Computation}. We adopt Alg514 \cite{kushman2014learning} and Dolphin18K \cite{huang2016well} for building English evaluation dataset. And we use Math23K \cite{wang2017deep} to build Chinese evaluation dataset. Each original instance consists of three parts, i.e., a question, its corresponding equation and answer. We use the answer to replace the wh-phrase in the question to construct a new input. Take the fifth example in Table \ref{tab:example} for example. The original question is ``\textit{Tony planted a 4 foot tree. The tree grows at a rate of 5 feet every year. How many years will it take to be 29 feet?}'' and the answer is ``5''. We only select simple questions whose equation has no more than two operators.

\paragraph{Masked Word Selection}
In each created input, we select an appropriate word or phrase (denoted as $w_m$) to mask. Then the sentence with a mask is input to a pre-trained LM, which will output a prediction for the masked position. Usually, we take $w_m$ as the golden standard answer for the masked position.
To precisely evaluate model prediction performance, the golden answer for the masked position in the context of the sentence should be as unique as possible. So the uniqueness of answer is one of our criteria for selecting masked words.

For the dimensions of ``knowledge'' and ``computation'', the answer for the original question is masked.
For the other three dimensions, annotators need to select appropriate masked words according to the uniqueness principle. 
In order to ensure the diversity of linguistic features in the data of grammar dimension, the masked words cover all parts-of-speech to test LMs acquisition of subject-verb agreement, pronoun case agreement, verb tense agreement, comparative/superlative adjectives and so on. %masked words, masked words for the dimension of \textit{grammar} should cover all POS tags and as much grammatical knowledge as possible, e.g., number of nouns and verbs, tense of verbs, and comparative form of adjectives in English.
Masked words in the data of semantics category cover concepts, properties, concept anaphors and so on. 
%For \textit{reasoning}, the representative word from premise, cause or effect is chosen to be masked.

Then for each masked position and the corresponding golden answer, three annotators rate their confidences on a 4-point scale by judging \textit{ whether the golden answer is unique} (1), \textit{among the top 3 predictions} (2), \textit{among the top 5 predictions} (3), or \textit{none of the above} (4).
The masked position is considered to be appropriate if the confidence of each annotator is no more than 2, i.e., its golden answer is unique or among the top 3 predictions.
%\yaozong{这里第一遍没反应过来是怎么打分的，第一是（1），（2），（3），（4）没有联想是对应的分数，第二个是没反应过来confidence分数是越小越好的}

\subsection{Iterative Rationale Annotation}
\label{ssec:rationale_annotation}
%证据标注原则，充分、简洁
Given an input with a masked segment and the golden answer for the segment, the annotators highlight important input tokens that support the prediction as the rationale. In our work, there are two rationale criteria used in the annotation process.

\paragraph{Rationale Criteria} 
As discussed in recent studies of natural language understanding tasks, a rationale should satisfy sufficiency, compactness and comprehensiveness \cite{lei2016rationalizing, yu2019rethinking}. 
%where the \textcolor{yellow}{comprehensiveness requires all rationales to be selected, not just a sufficient set}. 
As the comprehensiveness is not suitable for the rationale of LMs' prediction, we use sufficiency and compactness as the rationale criteria.
\begin{itemize}[leftmargin=*]
\item \textbf{Sufficiency}. A rationale is sufficient if it contains enough information for people to make the correct prediction. In other words, human can make the correct prediction only based on tokens in the rationale. 
\item \textbf{Compactness}. A rationale is compact if all of its tokens are indeed required in making a correct prediction. That is to say, when any token is removed from the rationale, the prediction will change or become difficult to make.
%\item \textbf{Comprehensiveness}. A rationale is comprehensive if its complements in the input can not imply the prediction, that is, all evidence that supports the output should be labeled as rationales.
\end{itemize}

\paragraph{Annotation Process}
To ensure the data quality, we adopt an iterative annotation workflow, including three steps.

\textbf{Step 1: rationale annotation}. Given the input and the corresponding golden answer, the ordinary annotators label all critical tokens that are needed for correct prediction based on their intuition on the model decision mechanism. 
%as shown in Table \ref{tab:example}.

\textbf{Step 2: rationale scoring}. Our senior annotators double-check the annotations according to the annotation criteria. For each rationale, the annotators rate their confidences for sufficiency by judging \textit{whether they are unable} (1), \textit{probably able} (2), or \textit{definitely able} (3) to make the correct prediction only based on it, and rate their confidences for compactness by judging \textit{whether it contains redundant tokens} (1), \textit{contains disturbances} (2), \textit{is probably concise} (3), or \textit{is very concise} (4).%and compactness
%The confidences for \textbf{sufficiency} consist of three classes: \textit{can not support result (1)}, \textit{not sure (2)} and \textit{can support result (3)}. 
%And the confidences for \textbf{compactness} compose of four classes: \textit{include redundant tokens (1)}, \textit{include disturbances (2)}, \textit{not sure (3)} and \textit{conciseness (4)}. 

A rationale is considered to be of high-quality if its average score on sufficiency and compactness is equal to or greater than $3$ and $3.6$, respectively. All unqualified data whose average score on a property is lower than the corresponding threshold goes to the next step.

\textbf{Step 3: rationale modification}. Low-quality rationales are given to the ordinary annotators for correction.

Then the corrected rationales are scored by senior annotators again. This iterative annotation-scoring process runs for 3 iterations and the unqualified data is discarded after that.  

%是否介绍其他细节，如标注者信息，标注培训等
%Other annotation details, such as annotator information, annotation training and data usage instructions, are described in Appendix \ref{sec:data_other}.

\subsection{Perturbed Data Creation}
\label{ssec:disturb_data}
%重要term修改、不重要term修改、句式改变，原则都是不修改答案，除了计算类中，修改了一些数字，可能会修改答案

Recent studies \cite{ding2021evaluating, wang2021dutrust} propose to evaluate the model faithfulness via measuring how consistent its rationales are regarding perturbations. That is to say, under perturbations that are not supposed to change the model prediction mechanism, a model is considered faithful if its rationales are unchanged. 
In our work, we adopt this metric to evaluate LM's interpretability. And we construct perturbed examples for each original input.

\paragraph{Perturbation Criteria} 
%Perturbations aim to evaluate the faithfulness both for models and saliency methods. 
In our work, perturbations do not change the model prediction and internal decision mechanism. 
Please note that the influence of perturbations on model's prediction and decision mechanism comes from human's basic intuition. 
%We create perturbed examples considering two aspects: \textcolor{yellow}{1) perturbations should not influence model rationales; 2) perturbations can cause the alterations of rationales. }
%\yaozong{这里是指扰动会导致原本的证据变得不重要了，所以不会再被选为证据，还是说在证据文本上修改perturbations on rationales？}. 
Based on the literature \cite{jia2017adversarial, mccoy2019right, ribeiro2020beyond}, we define three perturbation types.
\begin{itemize}[leftmargin=*]
\item \textbf{Alteration of dispensable words (\textit{Dispens.})}, that is, inserting, deleting or replacing words that should have no effect on model predictions and rationales, e.g., the sentence ``the man's eyes were stabbed by broken glass, then he went blind'' is changed to ``unfortunately, the man's eyes were stabbed by broken glass, then he went blind''.
\item \textbf{Alteration of important words (\textit{Import.})}, that is, replacing important words which have an impact on model predictions with their synonyms or related words, for example, replacing ``stabbed'' with ``pierced''. In this situation, the rationale changes, but the prediction does not change. 
\item \textbf{Syntactic transformation (\textit{Trans.})}, transforming the syntactic structure of an instance without changing its meaning, e.g., the voice change from ``the man's eyes were stabbed by broken glass'' to ``the broken glass stabbed the man's eyes''. In this case, the model prediction and rationale should not be affected. 
\end{itemize}

We create at least one perturbed example for each original input. And we annotate at least 100 perturbed examples for each perturbation type. 
We ask two annotators to create perturbed examples, and ask two other annotators to review and modify the created examples.

\subsection{Data Statistics}
\label{ssec:data_stas}

Table \ref{tab:data_stas} shows the detailed statistics of our evaluation benchmark. We can see that the number of pairs and the length ratio of rationale vary with evaluation dimensions. 
As discussed above (i.e., ``Masked word selection'' in Section \ref{ssec:data_collection}), the instances in the dimension of grammar cover as much syntactic knowledge as possible. The English grammar dataset is larger in size than Chinese ones as there are less agreement rules in Chinese grammar. 
The Chinese dataset for semantics is larger than the English ones as there are more available data in Chinese. 
The rationale length ratio can be used as the reference length of the rationales during extracting. And the rationale length affects interpretability performance, as discussed in Section \ref{ssec:result_analysis}. 

\begin{table}[tb]
\renewcommand\tabcolsep{2.5pt}
\centering
\begin{tabular}{l | r r | r r}
\toprule
\multirow{2}{*}{Dimensionality} & \multicolumn{2}{c|}{English} & \multicolumn{2}{c}{Chinese} \\
\cline{2-3} \cline{4-5}
 & Size & RRL(\%) & Size & RRL(\%) \\
\hline
Grammar & 1,365 & 29.8 & 701 & 20.7 \\
Semantics & 793 & 31.6 & 1,210 & 27.1 \\
Knowledge & 295 & 45.8 & 300 & 51.5 \\
Reasoning & 300 & 48.5 & 300 & 43.4 \\
Computation & 307 & 59.7 & 400 & 54.5 \\
\bottomrule
\end{tabular}
\caption{Statistics of our datasets. ``Size'' means the number of original/perturbed pairs. ``RRL'' represents the ratio of rationale length to its input length.}
\label{tab:data_stas}
\end{table} 

\section{Metrics}
\label{sec:metrics}
Following previous works \cite{arras2017relevant, mohseni2018human, weerts2019human, atanasova2020diagnostic, liu2020interpretations}, we evaluate interpretability from the aspects of plausibility and faithfulness. 
%In this section, we introduce the metrics that are used in our work.

\paragraph{Plausibility} 
Plausibility measures how well the rationale provided by the model aligns with the human-annotated rationale \cite{jacovi2020towards, deyoung2020eraser, ding2021evaluating}. We adopt token F1-score as the metric for plausibility, as shown in Equation \ref{equation:marco-f1}.
For each model prediction, we select the top $K$ important tokens to compose its rationale, where the token importance score is assigned by a specific saliency method. In our experiments, $K$ is the product of the average rationale length ratio $r$ and the current input length $l$.
\begin{equation}
\small
\centering
\begin{aligned}
\texttt{Token-F1} &= \frac{1}{N}\sum_{i=1}^N (2 \times \frac{P_i \times R_i}{P_i+R_i}) \\
\texttt{where} \quad P_i&=\frac{|S_i^p \cap S_i^g|}{|S_i^p|} \,\, \texttt{and} \,\, R_i=\frac{|S_i^p \cap S_i^g|}{|S_i^g|}
\end{aligned}
\label{equation:marco-f1}
\end{equation}
where $S_i^p$ and $S_i^g$ represent the model's rationale and human-annotated rationale of the $i$-th instance, respectively; $N$ represents the total number of instances.

\paragraph{Faithfulness}
Faithfulness evaluates to what extent the rationale provided by the model truly affects the model prediction \cite{jacovi2020towards, ding2021evaluating}. 
A variety of metrics have been proposed to evaluate faithfulness from multiple perspectives, e.g. sufficiency and comprehensiveness \cite{deyoung2020eraser}, consistency under perturbations \cite{ding2021evaluating, wang2021dutrust}, sensitivity and stability \cite{yin2021on}. Most of these evaluation metrics are only applicable to classification models.
%and word-level differential reaction \cite{mosca2022that}.

Considering the characteristics of pre-trained LMs, we evaluate faithfulness for pre-trained LMs via evaluating the consistency of rationales under perturbations. In our work, we adopt Mean Average Precision (MAP) \cite{wang2021dutrust} and Pearson Correlation Coefficient (PCC) \cite{ding2021evaluating} to evaluate the consistency of two rationales. 

\textbf{MAP}, as defined in Equation \ref{equation:map}, evaluates the consistency of rationales under perturbations by calculating the order consistency of two token lists, i.e., the token list of the original instance and that of the corresponding perturbed instance. The higher the MAP, the more faithful the rationale.
\begin{equation}
\small
\centering
\texttt{MAP} = \frac{\sum_{i=1}^{|X^p|}(\sum_{j=1}^i G(x_j^p, X_{1:i}^o))/i}{|X^p|}
\label{equation:map}
\end{equation}
where $X^o$ and $X^p$ are the sorted token lists of the original and perturbed inputs respectively. $|X^p|$ represents the token number of $X^p$. $X_{1:i}^o$ contains the top-$i$ important tokens of $X^o$. The function $G(x,Y)$ determines whether the token $x$ belongs to the list $Y$. In other words, $G(x,Y)=1 \, \texttt{iff} \, x \in Y$. 

\textbf{PCC} is the test statistics that measures the statistical association between two continuous variables. As shown in Equation \ref{equation:PCC}, we use it to measure the linear correlation between token importance scores of the original instance and that of the perturbed one. Based on perturbation types defined in Section \ref{ssec:disturb_data}, we post process the two importance score lists to align them. Specifically, the unaligned tokens, such as the deleted words and inserted words, will be aligned to a virtual token whose importance score is 0. 
As it is difficult to align the perturbed instance with its original instance under the perturbation type of \textit{Trans.}, we do not perform PCC calculation on pairs of \textit{Trans.} type.
The PCC value ranges from 0 to 1. The higher the PCC score\footnote{In our experiments, our reported PCC values are computed on pairs with $\texttt{p-value}<0.05$, where p-value represents the significance level of the linear correlation.}, the more faithful the rationale. 
%To compare the PCC values with MAP values, MAP values on the selected data are also calculated. 放在实验部分
\begin{equation}
\small
\centering
  \texttt{PCC} =
  \frac{ \sum_{i=1}^{N}(v_i^o-\bar{v}^o)(v_i^p-\bar{v}^p) }{%
        \sqrt{\sum_{i=1}^{N}(v_i^o-\bar{v}^o)^2}\sqrt{\sum_{i=1}^{N}(v_i^p-\bar{v}^p)^2}}
  \label{equation:PCC}
\end{equation}
where $v_i^o$ and $v_i^p$ represent the $i$-th elements of importance score vectors of the original and perturbed instance, respectively. $\bar{v}^o$ and $\bar{v}^p$ are the mean of $v^o$ and $v^p$ respectively.

From the definitions of MAP and PCC, it can be seen that MAP measures the association of two token lists based on importance order, and PCC assesses the association of two token lists according to their importance values.
\section{Experiments}
\label{sec:experiments}
\begin{table*}[]
\renewcommand\tabcolsep{2.5pt}
\centering
\scalebox{0.9}{
\begin{tabular}{l|ccc|ccc|ccc|ccc|ccc}
\toprule
\multirow{2}{*}{ Model + TopN} & \multicolumn{3}{c|}{Grammar} & \multicolumn{3}{c|}{Semantics} & \multicolumn{3}{c|}{Knowledge} & \multicolumn{3}{c|}{Reasoning} & \multicolumn{3}{c}{Computation} \\ 
\cline{2-16} 
 & All & Ori. & Per. & All & Ori. & Per. & All & Ori. & Per.  & All & Ori. & Per. & All & Ori. & Per. \\ 
\hline
%BERT-base + Top1 & 61.27 & 61.34 & 61.20 & 43.87 & 42.43 & 45.32 & 7.17  & 8.33 & 6.00 & 20.50 & 21.67 & 19.33 & 1.12 & 1.50 & 0.75 \\ 
%BERT-base + Top1 & 61.3 & 61.3 & 61.2 & 43.9 & 42.4 & 45.3 & 7.2  & 8.3 & 6.0 & 20.5 & 21.7 & 19.3 & 1.1 & 1.5 & 0.8 \\ 
BERT-base + Top1 & 61.3 & 61.3 & 61.2 & 43.9 & 45.3 &42.4 &  7.2  & 8.3 & 6.0 & 20.5 & 21.7 & 19.3 & 1.1 & 1.5 & 0.8 \\ 
%BERT-base + Top3 & 79.24 & 80.31 & 78.17 & 61.70 & 60.21 & 63.18 & 12.17 & \textbf{13.00} & 11.33 & 32.83 & 35.00 & 30.67 & \textbf{3.25} & \textbf{4.00} & 2.50 \\ 
%BERT-base + Top3 & 79.2 & 80.3 & 78.2 & 61.7 & 60.2 & 63.2 & 12.2 & \textbf{13.0} & 11.3 & 32.8 & 35.0 & 30.7 & \textbf{3.3} & \textbf{4.0} & 2.5 \\ 
BERT-base + Top3 & 79.2 & 80.3 & 78.2 & 61.7 & 63.2& 60.2  & 12.2 & \textbf{13.0} & 11.3 & 32.8 & 35.0 & 30.7 & \textbf{3.3} & \textbf{4.0} & 2.5 \\ 
\hline
%RoBERTa-base + Top1 & 67.55 & 68.76 & 66.33 & 51.03 & 48.93 & 53.14 & 5.00 & 5.33 & 4.67 & 23.80 & 25.00 & 22.67 & 1.00 & 0.75 & 1.25 \\ 
%RoBERTa-base + Top3 & 81.67 & 82.17 & 81.17 & 71.90 & 71.40 & 72.40 & 10.00 & 8.67 & 11.33 & 43.50 & 43.00 & 44.00 & 2.88 & 3.25 & 2.50 \\ 
%RoBERTa-base + Top1 & 67.6 & 68.8 & 66.3 & 51.0 & 48.9 & 53.1 & 5.0 & 5.3 & 4.7 & 23.8 & 25.0 & 22.7 & 1.0 & 0.8 & 1.3 \\ 
%RoBERTa-base + Top3 & 81.7 & 82.2 & 81.2 & 71.9 & 71.4 & 72.4 & 10.0 & 8.7 & 11.3 & 43.5 & 43.0 & 44.0 & 2.9 & 3.3 & 2.5 \\ 
RoBERTa-base + Top1 & 67.6 & 68.8 & 66.3 & 51.0 & 53.1& 48.9  & 5.0 & 5.3 & 4.7 & 23.8 & 25.0 & 22.7 & 1.0 & 0.8 & 1.3 \\ 
RoBERTa-base + Top3 & 81.7 & 82.2 & 81.2 & 71.9 & 72.4& 71.4  & 10.0 & 8.7 & 11.3 & 43.5 & 43.0 & 44.0 & 2.9 & 3.3 & 2.5 \\ 
\hline
%ERNIE-base + Top1 & 64.27 & 65.48 & 63.05 & 50.62 & 49.67 & 51.57 & 4.50 & 4.00 & 5.00 & 25.00 & 26.67 & 23.33 & 0.25 & 0.25 & 0.25 \\ 
%ERNIE-base + Top3 & 81.74 & 82.60 & 80.88 & 71.69 & 71.08 & 72.31 & \textbf{12.33} & 12.00 & \textbf{12.67} & 45.50 & 47.00 & 44.00 & 1.62 & 1.50 & 1.75 \\ 
%ERNIE-base + Top1 & 64.3 & 65.5 & 63.1 & 50.6 & 49.7 & 51.6 & 4.5 & 4.0 & 5.0 & 25.0 & 26.7 & 23.3 & 0.3 & 0.3 & 0.3 \\ 
%ERNIE-base + Top3 & 81.7 & 82.6 & 80.9 & 71.7 & 71.1 & 72.3 & \textbf{12.3} & 12.0 & \textbf{12.7} & 45.5 & 47.0 & 44.0 & 1.6 & 1.5 & 1.8 \\ 
ERNIE-base + Top1 & 64.3 & 65.5 & 63.1 & 50.6 & 51.6& 49.7  & 4.5 & 4.0 & 5.0 & 25.0 & 26.7 & 23.3 & 0.3 & 0.3 & 0.3 \\ 
ERNIE-base + Top3 & 81.7 & 82.6 & 80.9 & 71.7 & 72.3& 71.1  & \textbf{12.3} & 12.0 & \textbf{12.7} & 45.5 & 47.0 & 44.0 & 1.6 & 1.5 & 1.8 \\ 
\hline
%ERNIE-large + Top1 & 67.69 & 68.47 & 66.90 & 52.98 & 51.57 & 54.38 & 4.33  & 4.00  & 4.67 & 30.50 & 32.00 & 29.00 & 0.50 & 0.25 & 0.75 \\ 
%ERNIE-large + Top3 & \textbf{83.95} & \textbf{84.88} & \textbf{83.02} & \textbf{73.59} & \textbf{72.56} & \textbf{74.63} & 11.17 & 10.33 & 12.00 & \textbf{50.17} & \textbf{52.67} & \textbf{47.67} & 2.75 & 2.50 & \textbf{3.00} \\ 
%ERNIE-large + Top1 & 67.7 & 68.5 & 66.9 & 53.0 & 51.6 & 54.4 & 4.3 & 4.0 & 4.7 & 30.5 & 32.0 & 29.0 & 0.5 & 0.3 & 0.8 \\ 
%ERNIE-large + Top3 & \textbf{84.0} & \textbf{84.9} & \textbf{83.0} & \textbf{73.6} & \textbf{72.6} & \textbf{74.6} & 11.2 & 10.3 & 12.0 & \textbf{50.2} & \textbf{52.7} & \textbf{47.7} & 2.8 & 2.5 & \textbf{3.0} \\ 
ERNIE-large + Top1 & 67.7 & 68.5 & 66.9 & 53.0 & 54.4& 51.6  & 4.3 & 4.0 & 4.7 & 30.5 & 32.0 & 29.0 & 0.5 & 0.3 & 0.8 \\ 
ERNIE-large + Top3 & \textbf{84.0} & \textbf{84.9} & \textbf{83.0} & \textbf{73.6} & \textbf{74.6}& \textbf{72.6}  & 11.2 & 10.3 & 12.0 & \textbf{50.2} & \textbf{52.7} & \textbf{47.7} & 2.8 & 2.5 & \textbf{3.0} \\ 
\bottomrule
\end{tabular}
}
\caption{Model performance on masked word predictions, where \textit{Ori.} and \textit{Per.} represent performance on original inputs and perturbed inputs respectively, and \textit{All} represents performance on all inputs.}
\label{tab:model-ability-with-ori-dis}
\end{table*}
\begin{table*}[]
\renewcommand\tabcolsep{2.5pt}
\centering
\scalebox{0.68}{
\begin{tabular}{l | c c c | c c c | c c c | c c c | c c c}
\toprule
\multirow{2}{*}{\makecell[c]{Model + Method}} & \multicolumn{3}{c|}{Grammar} & \multicolumn{3}{c|}{Semantics} & \multicolumn{3}{c|}{Knowledge} & \multicolumn{3}{c|}{Reasoning} & \multicolumn{3}{c}{Computation}  \\ 
\cline{2-16}
%\hline
 & F1 & MAP & PCC/MAP$^*$ & F1 & MAP & PCC/MAP$^*$ & F1 & MAP & PCC/MAP$^*$ & F1 & MAP & PCC/MAP$^*$ & F1 & MAP & PCC/MAP$^*$ \\ 
\hline
%BERT-base + ATT & 34.69 & 85.24 & 38.58 & 81.42 & 67.65 & 73.42 & 52.93 & 72.54 & 63.83 & 83.20\\
%BERT-base + IG & 34.88 & 69.79 & 29.81 & 67.26 & 60.92 & 58.99& 48.69 & 55.44 & 59.27 & 75.96 \\
%BERT-base + ATT & 0.35 & 0.85 & 0.93 / 0.91 & 0.39 & 0.81 & 0.94 / 0.91 & 0.68 & 0.73 & 0.92 / 0.90 & 0.53 & 0.73 & 0.88 / 0.84 & 0.64 & 0.83 & 0.87 / 0.89 \\
%BERT-base + IG & 0.35 & 0.70 & 0.88 / 0.73 & 0.30 & 0.67 & 0.89 / 0.75 & 0.61 & 0.59 & 0.83 / 0.71 & 0.49 & 0.55 & 0.80 / 0.63 & 0.59 & 0.76 & 0.88 / 0.78 \\
BERT-base + ATT & 0.35 & 0.85 & 0.93 / 0.90 & 0.39 & 0.82 & 0.94 / 0.91 & 0.68 & 0.73 & 0.92 / 0.90 & 0.53 & 0.73 & 0.88 / 0.84 & 0.64 & 0.83 & 0.87 / 0.89 \\
BERT-base + IG & 0.35 & 0.70 & 0.88 / 0.74 & 0.30 & 0.67 & 0.89 / 0.75 & 0.61 & 0.59 & 0.83 / 0.71 & 0.49 & 0.55 & 0.80 / 0.62 & 0.59 & 0.76 & 0.88 / 0.78 \\
\hline
%RoBERTa-base + ATT & 37.55 &	\textbf{86.05} & 37.00 &	\textbf{83.46} &	65.36 &	71.17  & 50.31 & 76.47&	63.69 & 85.21\\
%RoBERTa-base + IG & 30.28 & 72.34 & 31.19&	68.10&	62.54&	62.95&	49.80&	65.28&	59.84&	79.98 \\
%RoBERTa-base + ATT & 0.38 &	\textbf{0.86} & 0.93 / 0.91 & 0.37 & \textbf{0.83} & 0.94 / 0.92 &	0.65 & 0.71 & 0.92 / 0.90 & 0.50 & 0.76 & 0.90 / 0.86 & 0.64 & 0.85 & 0.89 / 0.90 \\
%RoBERTa-base + IG & 0.30 & 0.72 & 0.88 / 0.80 & 0.31 & 0.68 & 0.87 / 0.81 & 0.63 & 0.63 & 0.82 / 0.87 & 0.50 & 0.65 & 0.82 / 0.81 & 0.60 & 0.80 & 0.86 / 0.87 \\
RoBERTa-base + ATT & 0.38 &	\textbf{0.86} & 0.93 / 0.91 & 0.37 & \textbf{0.84} & 0.94 / \textbf{0.92} &	0.65 & 0.71 & 0.92 / 0.90 & 0.50 & 0.76 & 0.90 / 0.86 & 0.64 & 0.85 & 0.89 / 0.90 \\
RoBERTa-base + IG & 0.30 & 0.72 & 0.88 / 0.80 & 0.31 & 0.68 & 0.87 / 0.81 & 0.63 & 0.63 & 0.82 / 0.86 & 0.50 & 0.65 & 0.82 / 0.81 & 0.60 & 0.80 & 0.86 / 0.87 \\
\hline
%ERNIE-base + ATT & \textbf{50.29}  &  85.59 & \textbf{49.42}  &  81.86  & 70.72  &  74.21  & \textbf{65.54}  &  \textbf{78.47}  & \textbf{68.74}  &  \textbf{85.55}  \\
%ERNIE-base + IG & 37.71  &  65.91  & 36.40  &  61.58  & 63.67 &  54.33  & 54.09 & 53.31  & 61.35  &  75.99  \\
ERNIE-base + ATT & \textbf{0.50} & 0.86 & \textbf{0.95} / \textbf{0.91} & \textbf{0.49} & 0.82 & \textbf{0.96} / 0.91 & 0.71 & 0.74 & \textbf{0.96} / \textbf{0.92} & \textbf{0.66} & \textbf{0.78} & \textbf{0.95} / \textbf{0.88} & \textbf{0.69}  &  \textbf{0.86} & \textbf{0.94} / \textbf{0.91} \\
ERNIE-base + IG & 0.38 & 0.66 & 0.89 / 0.71 & 0.36 & 0.61 & 0.88 / 0.69 & 0.64 & 0.54 & 0.84 / 0.68 & 0.54 & 0.53 & 0.82 / 0.59 & 0.61 & 0.76 & 0.91 / 0.77 \\
%ERNIE-base + ATT & \textbf{0.50} & 0.86 & \textbf{0.95} / 0.91 & \textbf{0.49} & 0.82 & \textbf{0.96} / 0.91 & 0.71 & 0.74 & \textbf{0.96} / 0.92 & \textbf{0.66} & \textbf{0.78} & \textbf{0.95} / 0.88 & \textbf{0.69}  &  \textbf{0.86} & \textbf{0.94} / 0.91 \\
%ERNIE-base + IG & 0.38 & 0.66 & 0.89 / 0.71 & 0.36 & 0.62 & 0.88 / 0.69 & 0.64 & 0.54 & 0.84 / 0.68 & 0.54 & 0.53 & 0.82 / 0.59 & 0.61 & 0.76 & 0.91 / 0.77 \\
\hline
%ERNIE-large + ATT & 39.54  &  82.95  & 41.10  &  79.66  & \textbf{71.47}  &  \textbf{75.86}  & 53.95  &  69.02  & 66.40  &  83.42\\
%ERNIE-large + IG & 33.27  &  55.64  & 33.35  &  50.51  & 61.50  &  51.16  & 49.96  &  47.27  & 59.62  &  67.63  \\
ERNIE-large + ATT & 0.40 & 0.83 & 0.92 / 0.88 & 0.41 & 0.80 & 0.90 / 0.89 & \textbf{0.71} & \textbf{0.76} & 0.87 / 0.88 & 0.54 & 0.69 & 0.87 / 0.81 & 0.66 & 0.83 & 0.86 / 0.89 \\
ERNIE-large + IG & 0.33 & 0.56 & 0.70 / 0.64 & 0.33 & 0.51 & 0.69 / 0.64 & 0.62 & 0.51 & 0.76 / 0.70 & 0.50 & 0.47 & 0.67 / 0.61 & 0.60 & 0.68 & 0.72 / 0.73 \\
%ERNIE-large + ATT & 0.40 & 0.83 & 0.92 / 0.88 & 0.41 & 0.80 & 0.90 / 0.89 & \textbf{0.71} & \textbf{0.76} & 0.87 / 0.89 & 0.54 & 0.69 & 0.87 / 0.82 & 0.66 & 0.83 & 0.86 / 0.89 \\
%ERNIE-large + IG & 0.33 & 0.56 & 0.70 / 0.64 & 0.33 & 0.51 & 0.69 / 0.64 & 0.62 & 0.51 & 0.76 / 0.70 & 0.50 & 0.47 & 0.67 / 0.61 & 0.60 & 0.68 & 0.72 / 0.73 \\
\bottomrule
\end{tabular}
}
\caption{Interpretability evaluation of baseline LMs with two interpretation methods. As illustrated in Section \ref{sec:metrics}, the metric PCC is not performed on all inputs. For inputs suitable for PCC calculation, we compute MAP$^*$. }
\label{tab:result_summary_t}
\end{table*}

\subsection{Experimental Setting}
In order to evaluate model performance on our benchmark, we adopt several widely-used pre-trained LMs and interpretation methods as our baseline models and methods respectively. We only provide high-level descriptions for them and refer to the respective papers and source codes for details.

\paragraph{Evaluated Pre-trained LMs}
We conduct experiments on transformer-based pre-trained LMs. Specially, we adopt BERT-base-chinese (BERT-base) \cite{devlin2019bert}, RoBERTa-wwm-ext (RoBERTa-base) \cite{yiming2021roberta} and ERNIE (including ERNIE-base and ERNIE-large) \cite{sun2019ernie} as baseline models for Chinese\footnote{Since the parameters in the masked-LM layers of BERT-large and RoBERTa-large are not released with models, which are required in our experiments, we do not conduct experiments on large versions of BERT and RoBERTa.}. 
Meanwhile, we take BERT \cite{devlin2019bert} and RoBERTa \cite{liu2019roberta} as baseline models for English.
%Please note that we have not conducted experiment on large versions of some LMs, since the parameters of masked-LM layers are not released with models, which are required in our evaluation.

\paragraph{Interpretation Methods} 
We use attention (ATT) based method \cite{jain2019attention} and integrated gradient (IG) based method \cite{sundararajan2017axiomatic} for rationale extraction. For each input, we select the top-K important tokens to compose the rationale. In our experiments, K is the product of the average length ratio (i.e., $RRL$ in Table \ref{tab:data_stas}) and the current input length.

%放在附录中
In ATT based method, the attention weights in the last layer are taken as token importance scores. As the pre-trained LM uses wordpiece tokenization, we sum the self-attention weights assigned to its constituent pieces to compute a token's score. Meanwhile, we average scores over multi-heads to derive a final score. We denote this score as $S_{i,m}$ to represent the impact of token $i$ on predicting the masked token $m$. If the current prediction contains multiple masked tokens, the impact of each token is calculated by $S_{i,m} = \sum_{j\in{m}} S_{i,j}$, i.e., the sum of impact scores of all masked tokens. 

In IG based method, token importance is determined by integrating the gradient along the path from a defined baseline $x_0$ to the original input. In our experiments, the baseline $x_0$ is set as a sequence of ``[CLS] [PAD] ... [SEP]'', where the number of tokens of $x_0$ is equal to that of the original input. The step size is set to $100$. 

\paragraph{Evaluation Metrics}
We use a cloze-style task to evaluate model prediction performance, where a span of words in each input are replaced with ``[MASK]'' and the model need predict an answer for each ``[MASK]'' based on the context.
We evaluate model performance on the first (Top1) and the first three (Top3) predicted answers, using the prediction accuracy, i.e., the percentage of predictions that exactly match the golden answers. 

For interpretability evaluation, we use the metrics defined in Section \ref{sec:metrics}.

\subsection{Main Results}
\label{ssec:main_results}

In this section, we give an overview on model prediction performance and interpretability.

\paragraph{Model Prediction Performance}
Table \ref{tab:model-ability-with-ori-dis} shows model performance on masked word predictions.
It can be seen that all models perform well on instances of the types of grammar and semantics, which proves that pre-trained LMs have learned certain linguistic knowledge from large-scale corpus \cite{hewitt2019structural, jawahar2019does, tenney2019bert}.
However, on the other three evaluation dimensions, all models show a poor performance, especially on knowledge and computation. Existing studies also show that pre-trained LMs have no such abilities. For example, \citet{poerner2019bert} illustrate that BERT has not learned enough factual knowledge. \citet{hendrycks2021measuring, cobbe2021training} prove that pre-trained LMs can not handle math word problems.

%区别
\emph{Comparison between LMs}. From the comparisons between evaluated LMs, we get two interesting findings. First, RoBERTa and ERNIE perform better than BERT on dimensions of grammar, semantics and reasoning. Furthermore, ERNIE large outperforms ERNIE base on these three dimensions. We think there are two reasons, i.e., the larger size of training corpus and the larger size of parameters.
Second, BERT and ERNIE base have better performance on knowledge and computation. As discussed above, the abilities in these two dimensions have not been learned from the current training corpus and learning objectives. We think the relevant learning objectives need to be designed and the corresponding training data needs to be created.

\paragraph{Model Interpretability}
Table \ref{tab:result_summary_t} gives results on interpretability of different models and methods.
There are three main findings. 
Firstly, with two different interpretation methods, namely ATT and IG, all the evaluated LMs have a relatively strong faithfulness, which indicates that they are robust under perturbations. As shown in Table \ref{tab:model-ability-with-ori-dis}, compared with prediction accuracy on the original data, the prediction accuracy on perturbed data has not decreased too much. For example, in the dimension of grammar, prediction accuracy of most LMs is reduced by about 2\%. 
Secondly, across all evaluated LMs, ATT based method outperforms IG both in plausibility and faithfulness. We think this is because the interactions between words are more important for word generation based on the context. 
Thirdly, the metrics of plausibility (F1) and faithfulness (MAP) are positively correlated with the length ratio of extracted rationale (as shown in Table \ref{tab:data_stas}). Compared with performance on rationales, the performance on predictions is much poor. How to improve model prediction on plausible rationale is a problem that we should address in the future.

\emph{Comparison between LMs}. Compared ERNIE family with BERT and RoBERTa, ERNIE which is trained on a larger corpus performs better in plausibility with different interpretation methods. But RoBERTa has a higher MAP in the dimensions of grammar and semantics.
Compared ERNIE base and large, we find that the base-size model is superior to the large-size model on faithfulness and plausibility in all dimensions except for knowledge. This shows that larger parameter size may not lead to higher interpretability. 

\emph{Comparison between MAP and PCC}. As discussed in Section \ref{sec:metrics}, MAP measures faithfulness based on token importance order, while PCC relies on token importance values. From Table \ref{tab:result_summary_t}, we can see that the two metrics of the same model have the similar trend over different interpretation methods. However, the gap Between PCCs is smaller than that between MAPs.

% Side by side subfigures 
\begin{figure}
\begin{subfigure}[h]{0.2\linewidth}
\includegraphics[scale=0.4]{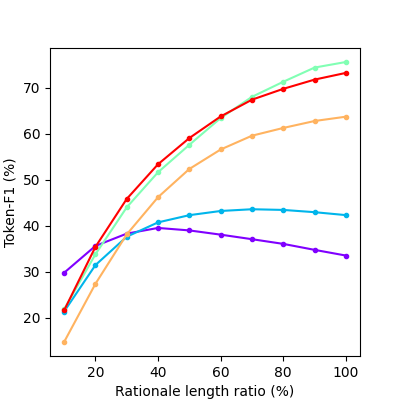}
\end{subfigure}
\hfill
\begin{subfigure}[h]{0.5\linewidth}
\includegraphics[scale=0.4]{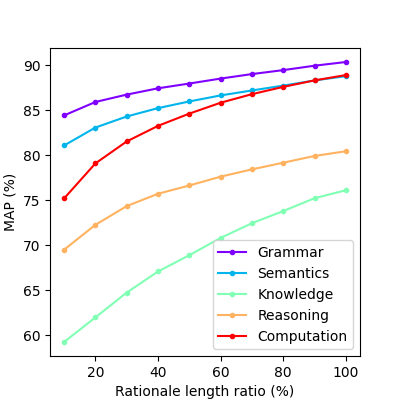}
\end{subfigure}%
\caption{Plausibility (F1) and faithfulness (MAP) of RoBERTa-base with ATT based interpretation method over different rationale length ratios.}
\end{figure}

\subsection{Analysis}
\label{ssec:result_analysis}

We give an in-depth analysis about prediction ability and interpretability according to the lengths of extracted rationales and perturbation types.

\paragraph{Analysis on Rationale Length}
We take the results of RoBERTa-base with ATT based method as an example to illustrate the impact of rationale length on interpretability, as shown in Figure \ref{fig:roberta-wwm-ext-att}. In the dimensions of knowledge, reasoning and computation, in which the length ratio of rationale is about 0.5, both plausibility and faithfulness increase with the increase of rationale length ratio. This proves that the most important words provided by the model and interpretation method perform poorly on interpretability. In the other two dimensions where the rationale length ratio is about 0.3, plausibility achieves the highest F1 score when the length ratio of extracted rationale is about 0.5. On the contrary, faithfulness (MAP) increases with the increase of rationale length. Compared with other three dimensions, MAP in these two dimensions increases slowly with the increase of rationale length.

\begin{table*}[]
\renewcommand\tabcolsep{2.5pt}
\centering
\scalebox{0.7}{
\begin{tabular}{l|c c c|c c c|c c c|c c c|c c c}
\toprule
\multirow{2}{*}{Data} & \multicolumn{3}{c|}{Grammar} & \multicolumn{3}{c|}{Semantics} & \multicolumn{3}{c|}{Knowledge} & \multicolumn{3}{c|}{Reasoning} & \multicolumn{3}{c}{Computation} \\
\cline{2-4} \cline{5-7} \cline{8-10} \cline{11-13} \cline{14-16}
& Dispens. & Import. & Trans. & Dispens. & Import. & Trans. & Dispens. & Import. & Trans. & Dispens. & Import. & Trans. & Dispens. & Import. & Trans. \\ 
\hline
%Original & 67.74 & 71.98  & 63.91 & 55.84 & 51.80 & 31.67 & 8.42 & 1.67 & 4.83 & 15.73 & 30.28 & 26.09 & 0.00 & 0.00 & 1.67 \\ 
%Perturbed & 66.45 (-1.29) & 70.33 (-1.65) & 60.15 (-3.76) & 57.51 (-1.67) & 51.54 (-0.26) & 46.97 (-15.3) & 6.32 (-2.1) & 1.67 (0.0) & 4.83 (0.0) & 15.73 (0.0) & 30.28 (0.0) & 15.94 (-10.15) & 1.64 (+1.64) & 1.67 (0.0) & 0.00 (0.1) \\ 
%Original & 67.7 & 72.0  & 63.9 & 55.8 & 51.8 & 31.7 & 8.4 & 1.7 & 4.8 & 15.7 & 30.3 & 26.1 & 0.0 & 0.0 & 1.7 \\ 
%Perturbed & \makecell[c]{66.4 \\ (-1.3)} & \makecell[c]{70.3 \\ (-1.7)} & \makecell[c]{60.1 \\ (-3.8)} & \makecell[c]{57.5 \\ (+1.7)} & \makecell[c]{51.5 \\ (-0.3)} & \makecell[c]{47.0 \\ (+15.3)} & \makecell[c]{6.3 \\ (-2.1)} & \makecell[c]{1.7 \\ (0.0)} & \makecell[c]{4.8 \\ (0.0)} & \makecell[c]{15.7 \\ (0.0)} & \makecell[c]{30.3 \\ (0.0)} & \makecell[c]{15.9 \\ (-10.2)} & \makecell[c]{1.6 \\ (+1.6)} & \makecell[c]{1.7 \\ (+1.7)} & \makecell[c]{0.0 \\ (-1.7)} \\ 
Original & 67.7 & 72.0  & 63.9 & 57.5 & 51.5 & 47.0 & 8.4 & 1.7 & 4.8 & 15.7 & 30.3 & 26.1 & 0.0 & 0.0 & 1.7 \\ 
Perturbed & \makecell[c]{66.4 \\ (-1.3)} & \makecell[c]{70.3 \\ (-1.7)} & \makecell[c]{60.1 \\ (-3.8)} & \makecell[c]{55.8 \\ (-1.7)} & \makecell[c]{51.8 \\ (+0.3)} & \makecell[c]{31.7 \\ (-15.3)} & \makecell[c]{6.3 \\ (-2.1)} & \makecell[c]{1.7 \\ (0.0)} & \makecell[c]{4.8 \\ (0.0)} & \makecell[c]{15.7 \\ (0.0)} & \makecell[c]{30.3 \\ (0.0)} & \makecell[c]{15.9 \\ (-10.2)} & \makecell[c]{1.6 \\ (+1.6)} & \makecell[c]{1.7 \\ (+1.7)} & \makecell[c]{0.0 \\ (-1.7)} \\ 
%influence(|dis - ori|)  & 1.65         & \textbf{3.76} & 1.29     & 0.26    & \textbf{15.3}  & 1.67     & 0.00       & 0.00             & \textbf{2.1} & 0.00       & \textbf{10.15} & 0.00        & \textbf{1.64} & 0.00             & 0.00        \\ \hline
%BERT-base-chinese + ori & 62.09        & 57.14         & 61.29    & 42.31   & 25.98          & 51.11    & 3.33    & 8.97          & 10.53        & 23.94   & 23.19          & 16.85    & 1.64          & 0.00             & 0.00        \\ 
%BERT-base-chinese + dis & 64.29        & 56.39         & 61.29    & 41.28   & 37.37          & 52.41    & 3.33    & 4.83          & 9.47         & 22.54   & 17.39          & 15.73    & 1.64          & 1.67          & 0.00        \\ 
%influence(|dis - ori|)  & \textbf{2.20} & 0.75          & 0.00        & 1.03    & \textbf{11.39} & 1.30      & 0.00       & \textbf{4.14} & 1.06         & 1.40     & \textbf{5.80}   & 1.12     & 0.00             & \textbf{1.67} & 0.00        \\ 
\bottomrule
\end{tabular}
}
\caption{Prediction accuracy of RoBERTa-base over different perturbation types.}
\label{tab:model-ability-adv-types}
\end{table*}
\begin{table*}[]
\renewcommand\tabcolsep{2.5pt}
\centering
\begin{tabular}{l|ccc|ccc|ccc}
\toprule
& \multicolumn{3}{c|}{Dispens.}                                                                                & \multicolumn{3}{c|}{Import.}                                                                                 & \multicolumn{3}{c}{Trans.}                                                                                  \\ \cline{2-10} 
\multirow{-2}{*}{Dimension} & F1$^o$ & F1$^p$ & MAP    & F1$^o$ & F1$^p$ & MAP    & F1$^o$ & F1$^p$ & MAP    \\ \hline
%Grammar & 36.79 & 36.65 & 90.64  & 38.66         & 37.59   & 83.38  & 37.75         & 37.85    & 81.02 \\ 
%Grammar & 0.37 & 0.37 & 0.90 & 0.96 / 0.93  & 0.39 & 0.38 & 0.83 & 0.89 / 0.88 &0.38 & 0.38 & 0.81 \\ 
Grammar & 0.370 & 0.367 & 0.906  & 0.387 & 0.376 & 0.834 &0.378 & 0.379 & 0.810 \\ 
%Semantic                                           & 37.54         & 36.74    & 88.10  & 39.39         & 39.81  & 86.76 & 32.38         & 33.83  & 69.97 \\ 
%Semantic & 0.38 & 0.37 & 0.88 & 0.95 / 0.92 & 0.39 & 0.40 & 0.87 & 0.92 / 0.91 & 0.32 & 0.34 & 0.70 \\ 
Semantics & 0.367 & 0.375 & 0.881 & 0.398 & 0.394 & 0.868  & 0.338 & 0.324 & 0.700 \\ 
%Knowledge                                          & 64.71         & 65.00      & 76.85 & 66.62         & 67.29   & 77.52  & 64.93         & 65.13    & 64.82 \\ 
%Knowledge & 0.65 & 0.65 & 0.77 & 0.91/0.89 & 0.67 & 0.67 & 0.78 & 0.94 / 0.91 & 0.65 & 0.65 & 0.65 \\ 
Knowledge & 0.647 & 0.650 & 0.769  & 0.666 & 0.673 & 0.775  & 0.649 & 0.651 & 0.648 \\ 
%Reasoning                                          & 51.56         & 52.82   & 80.93 & 48.33         & 50.92   & 78.62 & 48.28         & 50.31  & 66.29 \\ 
%Reasoning & 0.52 & 0.53 & 0.81 & 0.94 / 0.87 & 0.48 & 0.51 & 0.79 & 0.88 / 0.86 & 0.48 & 0.50 & 0.66 \\ 
Reasoning & 0.516 & 0.528 & 0.809  & 0.483 & 0.509 & 0.786  & 0.483 & 0.503 & 0.663 \\ 
%Computation                                        & 65.38         & 65.06   & 91.20  & 61.65         & 60.75   & 89.72 & 62.73         & 61.52& 83.32 \\ \bottomrule
%Computation & 0.65 & 0.65 & 0.91 & 0.96/0.94 & 0.62 & 0.61 & 0.90 & 0.95 / 0.94 & 0.63 & 0.62 & 0.83 \\ 
Computation & 0.654 & 0.651 & 0.912  & 0.617 & 0.608 & 0.897  & 0.627 & 0.615 & 0.833 \\ 
\bottomrule
\end{tabular}
\caption{Interpretability results of RoBERTa-base with attention based method under different perturbation types. F1$^o$ and F1$^p$ represents F1-scores on original and perturbed inputs respectively. }
\label{tab:interpretation-adv-types-fixed}
\end{table*}

\paragraph{Analysis on Perturbation Types}
In Table \ref{tab:model-ability-adv-types}, we give model prediction accuracy over three perturbation types. Due to space limitation, we take RoBERTa-base for example, as prediction accuracy of other LMs has the similar trend with RoBERTa-base. It can be seen that the prediction accuracy alters significantly on syntactically transformed perturbations (\textit{Trans.}). And the model is relatively robust on the other two perturbation types, i.e., alternation of dispensable words (\textit{Dispens.}) and alternation of important words (\textit{Import.}). 

Meanwhile, we further analyze interpretability results over different perturbation types, as shown in Table \ref{tab:interpretation-adv-types-fixed}. 
It can be seen that faithfulness (MAP) under \textit{Trans.} type is significantly lower than that under the other two perturbation types. 
Meanwhile, the \textit{Trans.} and \textit{Import.} types of perturbations have a larger influence on plausibility in the dimensions of semantics, reasoning and computation. 
%And \textit{Import.} influences plausibility in the dimension of knowledge.
Correspondingly, \textit{Dispens.} has little influence on interpretability, which proves that pre-trained LMs are robust to such perturbations.

\section{Conclusion}
To comprehensively evaluate pre-trained LMs, we construct a novel evaluation benchmark to evaluate both model prediction performance and interpretability from five perspectives, i.e., grammar, semantics, knowledge, reasoning and computation. 
%As far as we know, this is the first benchmark for model prediction and interpretability evaluation. 
We conduct experiments on several popular pre-trained LMs, and the results show that they perform very poorly in some dimensions, such as knowledge and computation. Meanwhile, the results show that their plausibility is far from satisfactory, especially when the rationale length ratio is small. Finally, the evaluated LMs have a strong robustness under perturbations, but they are less robust on syntax-aware data.
We will release this evaluation benchmark, and hope it will facilitate the research progress of pre-trained LMs.

%To better understand pre-trained LMs, we constructed five datasets in both Chinese and English. These datasets can be used to evaluate model ability and model interpretability. Our evaluation shows that current pre-trained LMs have a weak prediction ability in the dimensions of knowledge, reasoning, and computation. When it comes to model interpretability, attention based method has a better interpretation than integrated gradients method. On the same data dimension, the interpretation results shows similarity trend when increasing the rationale length ratio which indicates that the distribution of the importance scores is only related to tasks (i.e. data dimensions). Thus, the interpretation methods used have a stable explaining mechanism.
%未完待续

\bibliography{anthology,custom}

\clearpage
\appendix
\section{English results}
\label{sec:English_results}

\begin{table*}[]
\renewcommand\tabcolsep{2.5pt}
\centering
\scalebox{0.9}{
\begin{tabular}{l|ccc|ccc|ccc|ccc|ccc}
\toprule
\multirow{2}{*}{ Model + TopN} & \multicolumn{3}{c|}{Grammar} & \multicolumn{3}{c|}{Semantics} & \multicolumn{3}{c|}{Knowledge} & \multicolumn{3}{c|}{Reasoning} & \multicolumn{3}{c}{Computation} \\ 
\cline{2-16} 
 & All & Ori. & Per. & All & Ori. & Per. & All & Ori. & Per.  & All & Ori. & Per. & All & Ori. & Per. \\ 
\hline

BERT-base + Top1 & 52.2 & 52.5 & 51.9 & 64.7 & 65.6 & 63.8 & 0.7 & 0.7  & 0.7 & 24.0 & 23.7 & 24.3 & 0.5 & 0.3 & 0.7 \\
BERT-base + Top3 & 69.0 & 69.1 & 68.9 & 80.5 & 81.3 & 79.7 & 1.0 & 1.4  & 0.7 & 36.5 & 35.7 & 37.3 & 2.8 & 2.3 & 3.3 \\
\hline
BERT-large + Top1 & 58.8 & 58.9 & 58.8 & 67.9 & 69.1 & 66.7 & 0.7 & 1.0  & 0.3 & 28.7 & 28.0 & 29.3 & 1.3 & 1.0 & 1.6 \\
BERT-large + Top3 & 73.2 & 73.4 & 73.0 & 83.9 & 84.9 & 83.0 & 1.2 & 2.0  & 0.3 & 42.0 & 41.3 & 42.7 & 4.4 & 4.2 & 4.6 \\
\hline
RoBERTa-base + Top1 & 59.5 & 59.3 & 59.6 & 62.7 & 62.9 & 62.4 & 2.9 & 3.1  & 2.7 & 29.7 & 29.3 & 30.0 & 1.1 & 1.0 & 1.3 \\
RoBERTa-base + Top3 & 73.2 & 73.6 & 72.9 & 79.7 & 80.3 & 79.1 & 5.4 & 6.8  & 4.1 & 46.5 & 46.3 & 46.7 & 5.1 & 4.6 & 5.5 \\
\hline
RoBERTa-large + Top1 & 72.0 & 71.5 & 72.5 & 73.6 & 74.0 & 73.1 & 5.1 & 6.4  & 3.7 & 40.5 & 40.0 & 41.0 & 2.0 & 2.0 & 2.0 \\
RoBERTa-large + Top3 & \textbf{83.3} & \textbf{83.3}  & \textbf{83.4} & \textbf{89.0} & \textbf{89.5} & \textbf{88.4}  & \textbf{8.8} & \textbf{11.2} & \textbf{6.4}  & \textbf{57.8} & \textbf{59.7} & \textbf{56.0} & \textbf{6.7} & \textbf{7.2} & \textbf{6.2} \\ 
\bottomrule
\end{tabular}
}
\caption{Model performance on masked word predictions for English dataset, where \textit{Ori.} and \textit{Per.} represent performance on original inputs and perturbed inputs respectively, and \textit{All} represents performance on all inputs.}
\label{tab:model-ability-en-with-ori-dis}
\end{table*}
\begin{table*}[]
\renewcommand\tabcolsep{2.5pt}
\centering
\scalebox{0.68}{
\begin{tabular}{l | c c c | c c c | c c c | c c c | c c c}
\toprule
\multirow{2}{*}{\makecell[c]{Model + Method}} & \multicolumn{3}{c|}{Grammar} & \multicolumn{3}{c|}{Semantics} & \multicolumn{3}{c|}{Knowledge} & \multicolumn{3}{c|}{Reasoning} & \multicolumn{3}{c}{Computation}  \\ 
\cline{2-16}
%\hline
 & F1 & MAP & PCC/MAP$^*$ & F1 & MAP & PCC/MAP$^*$ & F1 & MAP & PCC/MAP$^*$ & F1 & MAP & PCC/MAP$^*$ & F1 & MAP & PCC/MAP$^*$ \\ 
\hline
BERT-base + ATT & 0.47 & \textbf{0.91}&\textbf{0.99} / \textbf{0.93}  & \textbf{0.44} & \textbf{0.84}&\textbf{0.97} / 0.90 & 0.59 & \textbf{0.73}&\textbf{0.99} / \textbf{0.85}  & 0.53 & \textbf{0.87}&\textbf{0.99} / \textbf{0.89} & 0.57 & \textbf{0.90}&\textbf{0.95} / \textbf{0.92} \\ 

BERT-base + IG  & 0.34 & 0.73 & 0.89 / 0.75  & 0.42  & 0.67&0.87 / 0.71 &  \textbf{0.66} & 0.51&0.83 / 0.63  & 0.56 & 0.64&0.83 / 0.68  & 0.62  & 0.79&0.85 / 0.80  \\ \hline

BERT-large + ATT & 0.47  &  0.87&0.95 / 0.91 & 0.40  &  0.83&0.94 / 0.89 & 0.58  &  0.64&0.94 / 0.85 & 0.59  &  0.84&0.96 / 0.87 & 0.60  &  0.90&0.94 / 0.90 \\

BERT-large + IG & 0.37  &  0.42& 0.45 / 0.48 & 0.41  &  0.38&0.44 / 0.46 & 0.64  &  0.40&0.54 / 0.61 & 0.57  &  0.41&0.39 / 0.51 & \textbf{0.66}  &  0.56&0.48 / 0.60 \\ \hline

RoBERTa-base + ATT & \textbf{0.55} & 0.88& 0.95 / 0.91   & 0.44  & 0.83&0.94 / 0.89  & 0.58  & 0.63&0.90 / 0.85  & \textbf{0.63}  & 0.83&0.92 / 0.88    & 0.62 & 0.87&0.91 / 0.90  \\ 

RoBERTa-base + IG  & 0.39 & 0.66& 0.78 / 0.73   & 0.37  & 0.56&0.72 / 0.68 & 0.56 & 0.41&0.73 / 0.67  & 0.56  & 0.59&0.79 / 0.69   & 0.63  & 0.73 &0.81 / 0.78\\ \hline

RoBERTa-large + ATT & 0.53  &  0.90&0.96 / 0.93 & 0.43  &  0.82&0.94 / \textbf{0.91} & 0.55  &  0.63&0.90 / 0.86 & 0.56  &  0.83&0.90 / 0.88 & 0.58  &  0.87&0.92 / 0.90 \\
RoBERTa-large + IG & 0.37  &  0.57&0.74 / 0.67 & 0.37  &  0.50&0.73 / 0.65 & 0.56  &  0.45&0.67 / 0.60 & 0.54  &  0.54&0.74 / 0.67 & 0.63  &  0.67&0.74 / 0.75 \\ \bottomrule
\end{tabular}
}
\caption{Interpretability evaluation of baseline LMs on English dataset with two interpretation methods. As illustrated in Section \ref{sec:metrics}, the metric PCC is not performed on all inputs. For inputs suitable for PCC calculation, we also compute MAP, denoted as MAP$^*$. }
\label{tab:result_summary_en}
\end{table*}

\begin{table*}[]
\renewcommand\tabcolsep{2.5pt}
\centering
\scalebox{0.7}{
\begin{tabular}{l|c c c|c c c|c c c|c c c|c c c}
\toprule
\multirow{2}{*}{Data} & \multicolumn{3}{c|}{Grammar} & \multicolumn{3}{c|}{Semantics} & \multicolumn{3}{c|}{Knowledge} & \multicolumn{3}{c|}{Reasoning} & \multicolumn{3}{c}{Computation} \\
\cline{2-4} \cline{5-7} \cline{8-10} \cline{11-13} \cline{14-16}
& Dispens. & Import. & Trans. & Dispens. & Import. & Trans. & Dispens. & Import. & Trans. & Dispens. & Import. & Trans. & Dispens. & Import. & Trans. \\ 
\hline
Original                                      & 59.8    & 58.6   & 54.3  & 60.2    & 73.9    & 63.2  & 4.3      & 1.6    & 3.1   & 25.6     & 30.0   & 35.5  & 2.6     & 1.1     & 0.0      \\ \hline
Perturbed                                     & \makecell[c]{60.4\\(+0.6)}     & \makecell[c]{59.0 \\ (+0.4)}   & \makecell[c]{51.2 \\ (-3.1)}  & \makecell[c]{59.2 \\ (-1.0)}    & \makecell[c]{71.8 \\ (-2.1)}   & \makecell[c]{64.6 \\ (+1.4)}  & \makecell[c]{2.9 \\ (-1.4)}      & \makecell[c]{4.9 \\ (+3.3)}    & \makecell[c]{1.8 \\ (-1.3)}   & \makecell[c]{28.1 \\ (+2.5)}     & \makecell[c]{30.5 \\ (+0.5)}   & \makecell[c]{32.3 \\ (-3.2)}  & \makecell[c]{2.6 \\ (0.0)}     & \makecell[c]{0.0 \\ (-1.1)}       & \makecell[c]{0.0 \\ (0.0)}      \\ \bottomrule
\end{tabular}
}
\caption{Prediction accuracy of RoBERTa-base over different perturbation types on English dataset.}
\label{tab:model-ability-en-adv-types}
\end{table*}
\begin{table*}[]
\renewcommand\tabcolsep{2.5pt}
\centering
\begin{tabular}{l|ccc|ccc|ccc}
\toprule
& \multicolumn{3}{c|}{Dispens.}                                                                                & \multicolumn{3}{c|}{Import.}                                                                                 & \multicolumn{3}{c}{Trans.}                                                                                  \\ \cline{2-10} 
\multirow{-2}{*}{Dimension} & F1$^o$ & F1$^p$ & MAP    & F1$^o$ & F1$^p$ & MAP    & F1$^o$ & F1$^p$ & MAP    \\ \hline
Grammar & 0.550 & 0.550 & 0.913   & 0.555 & 0.557 & 0.821  & 0.526 & 0.525 & 0.837    \\ 
Semantics & 0.395 & 0.396 & 0.854  & 0.543 & 0.542 & 0.856    & 0.448 & 0.432 & 0.644   \\ 
Knowledge & 0.575 & 0.559 & 0.783   & 0.597 & 0.614 & 0.750   & 0.601 & 0.544 & 0.512   \\
Reasoning & 0.624 & 0.622 & 0.874   & 0.627 & 0.625 & 0.821   & 0.629 & 0.616  & 0.753   \\ 
Computation & 0.624 & 0.623 & 0.921   & 0.613 & 0.617 & 0.877   & 0.585 & 0.604 & 0.888   \\
\bottomrule
\end{tabular}

\caption{Interpretability results of RoBERTa-base with attention based method over different perturbation types on English dataset. F1$^o$ and F1$^p$ represents F1-scores on original and perturbed inputs respectively. }
\label{tab:interpretation-adv-types-en}
\end{table*}

In this section, we show results on English dataset, as shown in Table \ref{tab:model-ability-en-with-ori-dis} - Table \ref{tab:interpretation-adv-types-en}.
Similarly, we give analyses from the perspectives of model prediction performance and interpretability. 
%The results on English dataset mostly supports the conclusions draw from Chinese dataset. However, there are several exceptions. Here, we analysis the difference between the experiment results from Chinese and English datasets on two aspects, i.e. model prediction ability and model interpretability.  

%模型能力与中文一致
\paragraph{Model Prediction Ability}
Table \ref{tab:model-ability-en-with-ori-dis} shows the prediction accuracy of evaluated LMs on English dataset. Generally, on English dataset, the performance on different dimensions has the similar trend with that on Chinese dataset. 
Firstly, all evaluated LMs perform very poorly on dimensions of knowledge and computation. 
Secondly, both for BERT and RoBERTa, the large-size model outperforms the base-size one. 
Thirdly, comparing models with the same size of parameters, RoBERTa which is trained on a larger corpus outperforms BERT on most of dimensions.

However, the robustness of the English LMs on different perturbation types is different from that of the Chinese LMs. As shown in Table \ref{tab:model-ability-en-adv-types}, RoBERTa-base is less robust under \textit{Trans.} compared with the other two perturbation types. But compared with the accuracy change under \textit{Trans.} on Chinese dataset, the accuracy change on English dataset is smaller. 

\paragraph{Model Interpretability}
Table \ref{tab:result_summary_en} shows the interpretation results of the evaluated LMs on the English dataset. 
Most of the conclusions on the Chinese dataset (illustrated in Section \ref{ssec:main_results}) are applicable to the English dataset.
However, on the English dataset, ATT based method not always performs better than IG based method on the two perspectives of interpretability. 
For example, on the dimensions of knowledge and computation, with all evaluated LMs, IG based method outperforms ATT based method on plausibility.

Meanwhile, Table \ref{tab:interpretation-adv-types-en} shows interpretability results of RoBERTa-base under different perturbation types. It can be seen that \textit{Trans.} brings a significant drop on faithfulness (MAP) on the dimensions of semantic, knowledge and reasoning. Correspondingly, on these three dimensions, plausibility on the perturbed data is lower than that on the orginal data. This proves that the evaluated LMs are less robust to \textit{Trans.} perturbation type.

%It can be seen that most conclusions form Section \ref{sec:experiments} can be consolidated with results on English dataset. However, there are two exceptions. Firstly, on English dataset, most evaluated models have a better plausibility on Knowledge and computation dimensions under IG method than under attention based method. Secondly, \textit{base} models do not always have a better plausibility than \textit{large} models as BERT-large have a better plausibility than BERT-base on grammar, reasoning and computation dimensions. 

%\emph{Comparison between perturbations.} Table \ref{tab:model-ability-en-adv-types} contains the prediction accuracy of RoBERTa-base over different perturbation types. Different than what we got on Chinese dataset, on English dataset, RoBERTa-base model shows its robustness under all three types of perturbations.
%Table \ref{tab:interpretation-adv-types-en} shows the interpretation results for RoBERTa-base with attention based method under different perturbation types. The evaluated metrics do not change much under different perturbations except ajie significant drop of the MAP value under \textit{Trans.} in semantic, knowledge and reasoning dimensions.

%以下表格如果使用，需要进一步修缮
%\input{tables/roberta-base-en-attention}
%\input{tables/roberta-base-en-ig}
%\input{tables/roberta-wwm-ext-ch-attention}
%\input{tables/roberta-wwm-ext-ch-ig}

\end{document}